# Interpretable AI plus Handheld, Portable Retinal Photographs: A Low-Cost Glaucoma Screening Solution for West Africa

**Charis Y. N. Chiang[1,2,3], Tarela Sarimiye[4], Adeyinka Ashaye[4], Martin Buist[2], Michael A. Hauser[5], Olusola Olawoye[4 *], Michaël J.A. Girard[1,3,6,7,8 *]**

1. Department of Ophthalmology, Emory University, Atlanta, Georgia, United States
2. Department of Biomedical Engineering, National University of Singapore, Singapore
3. Singapore Eye Research Institute, Singapore National Eye Centre, Singapore
4. Department of Ophthalmology, College of Medicine, University of Ibadan, Ibadan, Nigeria
5. Duke University School of Medicine, Durham, North Carolina, United States
6. Duke-NUS Graduate Medical School-Singapore, Singapore
7. Department of Biomedical Engineering, Georgia Institute of Technology, Atlanta, Georgia, United States
8. Emory Empathetic AI for Health Institute, Emory University, Atlanta, Georgia

* Both authors contributed equally and share senior and corresponding authorship.

**Word count:** 3674 (Manuscript Text)
256 (Abstract)

**Tables:** 2 + 2 (supplementary material)

**Figures:** 5 + 1 (supplementary material)

**Commercial relationship:** MJAG is the co-founder of the start-up company Abyss Processing Pte Ltd.

**Ethics statement:** This study was approved by the Institutional Ethics Review Board of the University of Ibadan/University College Hospital, Ibadan, Nigeria (UI/EC/24/0027), and all study procedures adhered to the tenets of the Declaration of Helsinki. Written informed consent was obtained from each subject.

**Support:** (1) the BrightFocus Foundation grant P383001572 (sub-award from Duke University) [MAH, OO]; (2) the Emory Eye Center (Emory University School of Medicine, Start-up funds, MJAG & CRE) [MJAG]; (3) a Challenge Grant from Research to Prevent Blindness, Inc. to the Department of Ophthalmology at Emory University [CYNC, MJAG]; (4) the NIH grant P30EY06360 to the Atlanta Vision Research Community [MJAG]; (5) the National Eye Institute of the National Institutes of Health under award numbers R01EY037299 and R01EY037245 [MJAG]; (6) the NMRC-LCG grant 'TAckling & Reducing Glaucoma Blindness with Emerging Technologies (TARGET)', award ID: MOH-OFLCG21jun-0003 [CYNC, MB, MJAG]; and (7) the NIH grant U01AG097746 [OO].

**Corresponding Authors:** Dr Michaël J.A. Girard
Ophthalmic Engineering & Innovation Laboratory (OEIL), Emory Eye Center, Emory School of Medicine, Emory Clinic Building B, 1365B Clifton Road, NE, Atlanta GA 30322
mgirard@ophthalmic.engineering
https://www.ophthalmic.engineering

Dr Olusola Olawoye
College of Medicine, University of Ibadan, Ibadan, Nigeria, +2348023890063
solaolawoye@yahoo.com

# ABSTRACT

**Purpose.** To develop and evaluate an interpretable artificial intelligence (AI) framework for glaucoma screening from low-cost portable, handheld retinal fundus photographs in a West African population and to compare its performance with clinical tabletop fundus imaging.

**Methods.** We used data from a community-based study of 681 participants (1,362 eyes) in Nigeria, comprising 414 glaucoma, 478 glaucoma suspect, and 470 non-glaucoma eyes. Fundus photographs were acquired using the low-cost handheld, portable Volk Viva retinal camera and the Canon CR-2-AF tabletop camera. We fine-tuned component models separately to each device to perform vessel segmentation, cup and disc boundary segmentation, and feature extraction to detect optic nerve head features. A final classification model combined these components to classify scans as glaucoma, glaucoma suspect or non-glaucoma. Feature-weight analysis and Gradient-weighted Class Activation Mapping were used for interpretation.

**Results.** The models performed well on both Volk Viva and Canon CR-2-AF images: Vessel segmentation: 0.98 Dice Coefficient (DC) (Volk) and 0.94 DC (Canon); Cup and disc segmentation: 0.95 DC (Volk) and 0.96 DC (Canon); Optic nerve head feature detection: area under the receiver operating characteristic curve (AUCs) of 0.83±0.03 (Volk) and 0.87±0.04 (Canon); Classification model: AUCs of 0.85±0.01 (Volk) and 0.93±0.01 (Canon). Reports for each image, present model decision confidence scores and decision-rationale visualizations to support clinical interpretation.

**Conclusions.** Volk Viva results were reasonably comparable to Canon CR-2-AF in the component models and not far behind in classification. This shows that interpretable AI combined with low-cost, portable imaging may enhance community-level glaucoma screening, especially in settings with limited specialist access and resources.

# INTRODUCTION

Glaucoma is the world's leading cause of permanent blindness [1], [2], characterized by distinct damage to the optic nerve [3], [4]. Early detection is crucial to slow disease progression and prevent irreversible vision loss [5]. This is particularly important in West African populations which bear a disproportionately high but understudied glaucoma burden [6], [7], [8]. However, limited access to eye care remains a major barrier to diagnosis. There is approximately one ophthalmologist per million population across West Africa, many without glaucoma-specific training, and very few glaucoma subspecialists [6], [7], [8]. Even in Nigeria, one of the better-resourced ophthalmic environments in West Africa, clinical need still far outstrips capacity. In Nigeria, glaucoma accounts for over 16% of blindness and affects around 5% of adults over 40 years old; however, only 12.5% of Nigerian ophthalmologists reported glaucoma subspeciality practice, and only 2.3% practice as strict glaucoma specialists [8].

In high-resource settings, glaucoma evaluation typically involves multiple tests. These include intraocular pressure measurement, visual field testing, optical coherence tomography, and fundus photography [9], [10]. Despite all this, glaucoma remains underdiagnosed even in resource-rich settings [11]. The high-cost clinical imaging systems, limited specialist availability, and limited clinical access amplify these barriers to glaucoma diagnosis in West Africa [7]. Therefore, scalable, low-cost screening approaches are urgently needed.

In recent years, portable fundus cameras have become increasingly affordable and portable. However, their utility in ophthalmology has been studied primarily in urban and clinical settings with few studies evaluating their use for glaucoma screening in low-resource community settings [12]. Studies conducted in Australian, American, and Nepalese clinical settings have reported mixed findings regarding the agreement between ophthalmologists' assessments of fundus photographs acquired using portable and conventional tabletop cameras [13], [14], [15], [16]. A separate hospital-based study in China found poor agreement (κ=0.32) between glaucoma predictions generated by the same deep learning algorithm from portable and tabletop fundus photographs [17]. In Nigeria, Garba et al. evaluated several portable tools in 524 eyes (5.92% glaucoma prevalence), including a Remidio portable fundus camera with integrated AI-based glaucoma detection, tonometry, visual-field testing, and visual acuity testing [18]. The portable fundus camera performed best, achieving an AUC of 0.91 for binary glaucoma classification [18]. However, this was a hospital-based study limited to binary glaucoma detection and there remains limited evidence for AI-based glaucoma screening from a single low-cost portable fundus photograph in West African community populations, particularly for distinguishing non-glaucoma, glaucoma suspect, and glaucoma in the absence of available glaucoma specialists. Additionally, although many glaucoma AI systems increasingly do incorporate explainability techniques such as class activation maps and cup and disc segmentation, relatively few provide a structured set of clinically interpretable intermediate findings that can be independently reviewed by the clinician alongside the final prediction [19], [20].

In this study, we developed an interpretable deep learning pipeline to screen for glaucoma from a color fundus photograph acquired using a low-cost portable, handheld camera. The pipeline automatically identified the optic cup, optic disc, retinal vessels, and twelve optic nerve head (ONH) features, and then integrated these outputs using a downstream AI classifier to categorize each eye as non-glaucomatous, glaucoma suspect, or glaucomatous. Finally, we applied explainability techniques to generate a report summarizing the image-derived findings and model rationale to support clinical interpretation of the results. This approach aims to support scalable glaucoma screening in West African populations where specialist access and clinical imaging resources are limited.

# METHODS

**Patient Recruitment**

A total of 681 subjects were included in this community-based study conducted in Nigeria. The cohort comprised 414 eyes with glaucoma, 478 glaucoma suspect eyes, and 470 non-glaucoma eyes. All subjects provided written informed consent. The study adhered to the tenets of the Declaration of Helsinki and was approved by the institutional Ethics review board of the University of Ibadan/University College Hospital Ibadan Nigeria (UI/EC/24/0027).

Diagnosis was performed by Nigerian clinicians with subspecialty training in glaucoma using anterior and posterior segment examination, dilated ONH assessment from slit lamp binocular fundus examination using a 78D lens, tabletop fundus photographs (Canon CR-2-AF), gonioscopy, visual field testing and optical coherence tomography where possible.

Glaucoma was defined as the presence of typical glaucomatous optic neuropathy, characterized by optic disc cupping, pallor, neuroretinal rim thinning, or retinal nerve fiber layer loss on optic disc examination and/or optical coherence tomography, with corresponding definite visual field loss. Although intraocular pressure was measured in all participants, it was not required for the diagnosis of glaucoma. Diagnostic classification followed the first two levels of evidence from the International Society of Geographical and Epidemiological Ophthalmology criteria. The first level required optic disc abnormality, defined as vertical cup-to-disc ratio (vCDR) ≥0.7, and/ or vCDR asymmetry ≥0.2, or focal neuroretinal rim narrowing to ≤0.1 cup-to-disc ratio (CDR) between 11–1 o'clock or 5–7 o'clock, together with a definite visual field defect consistent with glaucoma. The second level allowed a diagnosis of glaucoma in eyes with severe optic disc damage, defined as vCDR ≥0.9, when reliable visual field testing could not be performed due to poor vision.

Glaucoma suspects were defined as eyes with clinical findings suggestive of glaucoma that did not meet criteria for definite glaucoma. This group included optic disc suspects, visual field suspects, and ocular hypertension. Optic disc suspects had a glaucoma-appearing ONH, characterized by suspicious neuroretinal rim thinning, increased CDR, or inter-eye CDR asymmetry, without corresponding visual field loss. Visual field suspects had visual field defects suspicious for glaucoma without glaucomatous optic neuropathy. Ocular hypertension was defined as intraocular pressure >22 mmHg without glaucomatous optic neuropathy or visual field defects consistent with glaucoma.

Non-glaucomatous subjects were defined as eyes with intraocular pressure <22 mmHg and no evidence of glaucomatous optic neuropathy or other clinical evidence of glaucoma. Specifically, controls had no neuroretinal rim loss, and no definite visual field loss consistent with glaucomatous optic nerve damage.

**Fundus Photography**

Participants were recruited from community outreaches within Oyo State and imaged out of clinic, in the community, using a low-cost, handheld, portable Volk Viva retinal camera following pupil dilation to improve image clarity. The operators acquired between one and four photographs from each eye during the same imaging session using the Volk Viva camera. Where multiple Volk Viva photographs were acquired from the same eye, each photograph was retained as a separate image for image-level analyses. The same participants were then referred to the eye clinic of the University College Hospital Ibadan where they were imaged using a Canon CR-2-AF tabletop retinal camera and underwent other clinical tests. The respective cameras and sample photographs from each are shown in Figure 1.

Separate device-specific datasets were used for development and evaluation of the component models. The Volk Viva analysis cohort comprised 1,224 eyes from 630 participants after exclusion of corrupted images, while the Canon CR-2-AF cohort comprised 1,291 eyes from 665 participants as shown in Table 1. Component models were fine-tuned separately for each device to perform retinal vessel segmentation, cup and disc segmentation, and detection of ONH features. Images considered ungradable by CYNC because of poor image quality or inadequate visualization of the optic disc were excluded from analyses requiring these features. For development and comparison of the final diagnostic classifiers, the dataset was further restricted to eyes with gradable imaging available from both devices. This paired classification cohort comprised 1,091 eyes from 589 participants. The same paired cohort was used for both device-specific classifiers to permit direct comparison of their performance.

**Optic Nerve Head Feature Detector**

Previously, our group had manually reviewed a set of 566 fundus photographs, including photographs from publicly available datasets and a separate West African dataset [21], [22], [23], [24], [25], [26], [27], [28], [29], [30]. In each photograph, we determined whether the image quality was sufficient for grading. Images with insufficient quality for reliable grading, including poor focus, poor illumination, media opacity, or inadequate optic disc visibility, were classified as ungradable. We also determined whether each of the twelve fundus-derived ONH features were present or absent. These were, as depicted in Figure 2: tigroid fundus, saucerization, vessel bayonetting, visible lamina cribosa pores, notching, nasal peripapillary atrophy, temporal peripapillary atrophy, tortuous vessels, disc hemorrhage, bleeding, veinous occlusions, arterial occlusions. A ResNet18 model was then trained to automatically perform the grading. This pre-trained model was used as a starting point for the models trained in this study.

In this study, we manually reviewed and graded images from both the Volk Viva and Canon CR-2-AF datasets. Five-fold cross-validation was performed on 90% of the dataset, and a 10% hold-out set was used for testing. Identical subject-level partitions were used across the two imaging datasets to enable direct comparison between devices. All images from both eyes of the same participant were assigned to the same partition, ensuring no subject-level leakage across the training, validation, and test sets. Images were resized to 224 × 224 pixels and the previously trained model was finetuned to each dataset separately.

### Cup and Disc Segmentation Model

In previous work, cup and disc boundaries were manually segmented in 60 images from African American patients at Emory Hospital and 60 images from a separate dataset collected from West Africans. These annotations were used to train an initial UNET++ segmentation model to automatically delineate the optic cup and disc.

A custom graphical user interface was developed to manually annotate the optic cup and optic disc on fundus photographs. This was used to manually delineate the cup and disc in 60 Volk Viva and 60 Canon CR-2-AF fundus images. Our previously pretrained UNET++ model was then fine-tuned separately to the Volk Viva and Canon CR-2-AF fundus images.

All images were resized to 320 × 480 pixels before training. The model was trained using a Jaccard-index-based loss function averaged across segmentation classes. To reduce overfitting and improve generalizability, data augmentation was applied to the training set using the Albumentations Python library. Augmentations included horizontal flipping, random rotation and translation, additive Gaussian noise, and random changes in saturation and image intensity.

The segmentation dataset was split into training and test sets, with balanced representation of glaucomatous, suspect and non-glaucomatous eyes. Model performance was evaluated on held-out test images using the Dice coefficient, calculated by comparing predicted cup and disc masks with the corresponding manual annotations.

The cup and disc mask predictions were used to take vCDR and horizontal CDR (hCDR) measurements from each image.

### Vessel Segmentation Model

The vessels in the same 60 Volk Viva and 60 Canon CR-2-AF photos were manually annotated for training of a vessel segmentation model.

We pretrained another UNET++ segmentation model on 113 publicly available manually annotated fundus photographs from the Digital Retinal Images for Vessel Extraction dataset, Structured Analysis of the Retina dataset, Child Heart and Health Study in England database 1 dataset, and High-Resolution Fundus dataset [31], [32], [33], [34]. This pretrained UNET++ model was then separately fine-tuned to the manually annotated Volk Viva and Canon CR-2-AF photographs in a similar manner as the cup and disc segmentation models. The resulting vessel masks were used as an additional interpretable input for downstream glaucoma classification, capturing vascular patterns and vessel trajectories around the ONH.

### Classification Model

Glaucoma classification was performed as a three-class task distinguishing non-glaucoma, glaucoma suspects, and glaucoma eyes. Outputs from the component models were integrated into a downstream classification model to categorize each eye as non-glaucomatous, glaucoma suspect, or glaucomatous. Inputs included the original fundus image, cup and disc segmentation outputs, vessel segmentation masks, ONH feature predictions, and the participant’s age. The classifier was trained and evaluated separately for Volk Viva and Canon CR-2-AF images to compare performance of low-cost portable imaging with the gold standard of tabletop retinal photography.

The classification model was a late-fusion neural network comprising an ImageNet-pretrained ResNet18 backbone to process the raw colored fundus photograph, a convolutional branch for the cup-and-disc and vessel masks, and a tabular branch for all other variables. Five-fold cross-validation was performed on 90% of the dataset, and a 10% hold-out set was used for testing using the same splits as the feature-extraction model previously. Data augmentation was applied to the training set including horizontal flips, slight rotations, and Gaussian noise.

Model performance was evaluated using accuracy, class-balanced accuracy, class-specific precision and recall, and class-average one-vs-others area under the receiver operating characteristic curve (AUC).

A full overview of the automated pipeline is depicted in Figure 3.

**Interpretable Report**

For each image, the final pipeline generated an interpretable report summarizing the model outputs. This report included image quality, cup and disc segmentation, vessel segmentation, estimated cup-to-disc ratio, detected ONH features, predicted diagnostic class, and model confidence. Three complementary explainability methods were applied to better understand the model decision and generate clinically interpretable summaries to support review by eye care providers. One, ablation experiments were conducted by removing each input type, namely the photograph, masks, and features, in turn from the full model and measuring the resulting decline in AUC compared to the full classification model [35]; two, gradient-weighted class activation map (Grad-CAM) to identify the image regions contributing most strongly to the final classification [36]; and three, calculating Shapley Additive exPlanation (SHAP) values to rank tabular feature importance [37]. For the individual report, the absolute SHAP values were normalized across tabular features to express each feature's relative contribution as a percentage, such that the contributions summed to 100%.

# RESULTS

Of the 681 participants enrolled, Canon CR-2-AF photographs were acquired from 1,291 eyes of 665 participants, with one Canon photograph available per eye. A total of 2,777 Volk Viva photographs were acquired from 1,246 eyes of 642 participants. Forty-five Volk Viva images were corrupted on export, leaving 2,732 images from 1,224 eyes of 630 participants in the Volk Viva analysis cohort. Image-quality assessment identified 233 ungradable Volk Viva images and 113 ungradable Canon CR-2-AF images. Restricting the final diagnostic classification analysis to eyes with gradable imaging available from both devices resulted in a paired cohort of 1,091 eyes from 589 participants, comprising 287 glaucoma, 410 glaucoma suspect, and 394 non-glaucoma eyes as shown in Table 1.

The individual component models achieved excellent agreement with manual annotations for both retinal vessel and optic cup and disc segmentation (Table 2). For retinal vessel segmentation, Dice coefficients were 0.98 for Volk Viva images and 0.94 for Canon CR-2-AF images. Cup and disc segmentation similarly demonstrated high accuracy, with Dice coefficients of 0.95 and 0.96 for Volk Viva and Canon CR-2-AF images, respectively. Examples of the automated segmentations are shown in Figure 3. The feature extraction models achieved 0.83 ± 0.03 AUC and 0.87 ± 0.04 AUC for the Volk Viva and Canon CR-2-AF models, respectively.

Performance of the multimodal classification model differed between imaging devices, with the Canon CR-2-AF dataset achieving a macro-AUC of 0.93 ± 0.01 compared with 0.85 ± 0.01 for the portable Volk Viva dataset (Table 2). Receiver operating characteristic curves and confusion matrices are presented in Figure 4.

Ablation analysis revealed that the images and tabular features produced the largest reduction in classification performance for the Volk Viva classification model, while removal of segmentation masks contributed to the largest decrease in performance for the Canon CR-2-AF model (Supplementary Table 1). Across both datasets, SHAP analysis revealed that temporal peripapillary atrophy and CDR measurements were among the most influential in the tabular features, while tortuous vessels were also influential in the Canon CR-2-AF classification model (Supplementary Table 2). Grad-CAM analysis also revealed that the model generally focused on the disc region of each photo to make its decision (Supplementary Figure 1).

The model and explainability outputs were compiled to generate a report from each scan and a sample from a glaucomatous eye is shown in Figure 5.

# DISCUSSION

In this study, we developed an AI pipeline to screen for glaucoma specifically in a West African population. This was done using a single colored fundus photograph from a low-cost portable retinal camera. Rather than a black-box AI model, the pipeline was designed to parallel elements of clinical optic nerve assessment by first performing vessel segmentation, cup and disc segmentation and ONH feature detection before integrating these outputs with the fundus image in a downstream classifier to classify eyes into non-glaucoma, glaucoma suspect, and glaucoma. The pipeline also generated a clinician-friendly report. In this way, the screening aims to be transparent and accessible for clinicians working in a resource-limited setting where access to glaucoma specialists and advanced diagnostic equipment is scarce.

Overall, the proposed framework demonstrated good performance across each model component with comparable performance in both datasets. The vessel segmentation and cup-and-disc boundary delineation achieved strong agreement with manual annotations while the ONH feature detector accurately identified most of the visible features on images from both devices. These models performed similarly or even better than other AI applications in fundus photo datasets from other demographics [38], [39], [40], [41]. These results suggest that the modular approach, modelled after a clinician's mode of interpreting the same scans is feasible and can provide reliable intermediate outputs which clinicians can verify.

The performance achieved using the Canon CR-2-AF tabletop camera was only slightly better in classification performance than the low-cost Volk Viva portable retinal. Given the substantial difference in cost, portability, and ease of deployment between the two systems, this finding has important implications for expanding access to glaucoma screening in resource-limited settings. Consequently, affordable portable imaging devices can feasibly support community-based glaucoma screening and lighten the clinical workload

A key strength of this work is its transparency and interpretability. Although many deep learning systems have demonstrated high accuracy for glaucoma diagnosis and screening, most operate as black-box models that provide little insight into how a prediction is made [42], [43].

This lack of transparency may reduce clinician confidence and hinder implementation in routine clinical practice. Our framework addresses this limitation by generating an interpretable report comprising retinal vessel segmentation, cup and disc segmentation, and ONH feature detection alongside the final classification. This is particularly valuable in resource-limited settings where glaucoma specialists are scarce, and clinicians may benefit from rapid, structured interpretation of fundus photographs. Importantly, the intermediate outputs are themselves clinically meaningful, allowing clinicians to independently assess the ONH and draw their own conclusions without relying solely on the model's final classification. Furthermore, immediate classification of an image as ungradable would provide real-time feedback to the camera operator, allowing them to determine whether the image is suitable for report generation or whether an additional image should be acquired while still in the field.

The modality contribution results for the Volk Viva classification model indicate that the relative importance of image and tabular features was greater than the vessel and cup and disc masks, suggesting that though the segmentation masks did supplement the imaging data in leading to the model's classification decision, they did not dominate the decision. In comparison, the Canon CR-2-AF model showed substantially greater reliance on the masks. This may reflect the richer anatomical information available in the higher-quality Canon CR-2-AF images, including clearer cup and disc boundaries and a more extensive visible vascular network. Consequently, they may have contained more diagnostically informative structural information, whereas the sparser Volk Viva masks may have provided less information beyond that already captured by the RGB image and derived tabular features. However, it should be noted that because ablation measures the incremental contribution of each input in the presence of the remaining inputs, the small performance drop after removing Volk masks should not be interpreted as indicating that the masks contained no diagnostically relevant information. Amongst tabular features, SHAP analysis indicated that temporal peripapillary atrophy, and vertical and horizontal CDR were the most influential predictors. The Grad-CAM outputs also suggest that across both cameras and across diagnosis classes, the model focus was centred on the disc. Overall, these results indicate that structural optic disc measurements were important features for the multimodal classifier.

Several limitations should be discussed. First, the proposed pipeline consists of multiple models, meaning that errors from upstream tasks, such as image quality assessment, segmentation, or feature detection, may propagate to downstream classification performance and affect performance. Although each individual component demonstrated strong performance, diagnostic accuracy remains dependent on the reliability of each component model.

Second, the Canon CR-2-AF photographs were used as part of the clinical assessment that established the ground truth diagnosis. As such, there may be a methodological advantage for the Canon CR-2-AF dataset in comparison with the Volk Viva dataset. This may have contributed to the observed difference in classification performance between the two imaging devices.

Third, this was a single-cohort study conducted in a community-based population in Nigeria. External validation in independent cohorts of other West African cohorts is necessary before widespread deployment can be recommended.

Fourth, in many cases, the Canon CR-2-AF photos were sharper and thus contained more visible features than the Volk Viva photos. Although the Volk Viva camera achieved high performance for vessel segmentation, cup and disc segmentation, and ONH feature detection, these tasks were evaluated only on structures that were clearly visible in the images. Fine anatomical features, such as lamina cribrosa pores and subtle optic disc saucerization, and smaller vessels, were more readily visualized on Canon CR-2-AF images than the portable images. Features that could not be reliably observed were therefore excluded from the ground-truth annotations and were not evaluated by the detection model.

Finally, several ONH features such as notching, arterial and venous occlusion, and vessel bayonetting were uncommon or absent within the study population. This limited the amount of training data available for those individual feature detectors. Future studies involving larger and more diverse datasets may improve the detection of these relatively rare findings and further enhance the interpretability of the proposed framework.

In conclusion, we developed an interpretable deep learning framework capable of screening for glaucoma from a single portable fundus photograph while simultaneously providing clinically meaningful intermediate outputs, including cup and disc segmentation, vessel segmentation, and detection of ONH features. The comparable performance achieved using a low-cost portable camera demonstrates the potential for scalable glaucoma screening in underserved regions. With further external validation, this approach may facilitate earlier detection of glaucoma while providing clinicians with transparent, explainable decision support that is well suited for community-based eye care.

## ACKNOWLEDGEMENTS

We acknowledge support from **(1)** the BrightFocus Foundation grant P383001572 (sub-award from Duke University) [MAH, OO]; **(2)** the Emory Eye Center (Emory University School of Medicine, Start-up funds, MJAG & CRE) [MJAG]; **(3)** a Challenge Grant from Research to Prevent Blindness, Inc. to the Department of Ophthalmology at Emory University [CYNC, MJAG]; **(4)** the NIH grant P30EY06360 to the Atlanta Vision Research Community [MJAG]; **(5)** the National Eye Institute of the National Institutes of Health under award numbers R01EY037299 and R01EY037245 [MJAG]; **(6)** the NMRC-LCG grant 'TAckling & Reducing Glaucoma Blindness with Emerging Technologies (TARGET)', award ID: MOH-OFLCG21jun-0003 [CYNC, MB, MJAG]; and **(7)** the NIH grant U01AG097746 [OO].

**Table 1.** Cohort Demographics

| | **Volk Viva Dataset** | **Canon CR-2-AF Dataset** | **Paired Cohort for Classification Model** |
|---|---|---|---|
| No. subjects | 630 | 665 | 589 |
| Age | 52.3 ± 11.7 | 52.6 ± 12.0 | 51.4 ± 11.3 |
| Sex | 394 female; 236 male | 413 female; 252 male | 378 female; 211 male |
| No. Eyes | 1224 | 1291 | 1,091 |
| Glaucoma Eyes | 354 | 379 | 287 |
| Glaucoma Suspect Eyes | 440 | 466 | 410 |
| Non-Glaucoma Eyes | 430 | 446 | 394 |

**Table 2.** Performance Metrics from each Model

| Model | Volk Viva | Canon CR-2-AF |
|---|---|---|
| Vessel Segmentation | 0.98 Dice Coefficient | 0.94 Dice Coefficient |
| Cup/ Disc Segmentation | 0.95 Dice Coefficient | 0.96 Dice Coefficient |
| Feature Extraction | 0.83 ± 0.03 AUC | 0.87 ± 0.04 AUC |
| Classification | 0.85 ± 0.01 AUC | 0.93 ± 0.01 AUC |

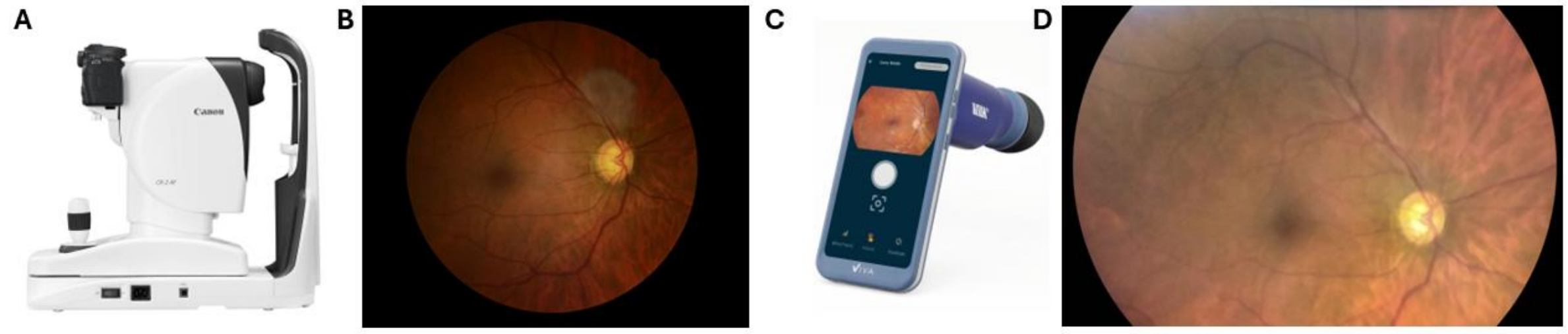

**Figure 1A**. Depiction of the Canon CR-2-AF tabletop camera; **B.** Fundus photo of a glaucomatous eye taken with Canon CR-2-AF tabletop camera. **C.** Depiction of the Volk Viva portable camera; **D.** Fundus photo of the same eye taken with Volk Viva portable camera.

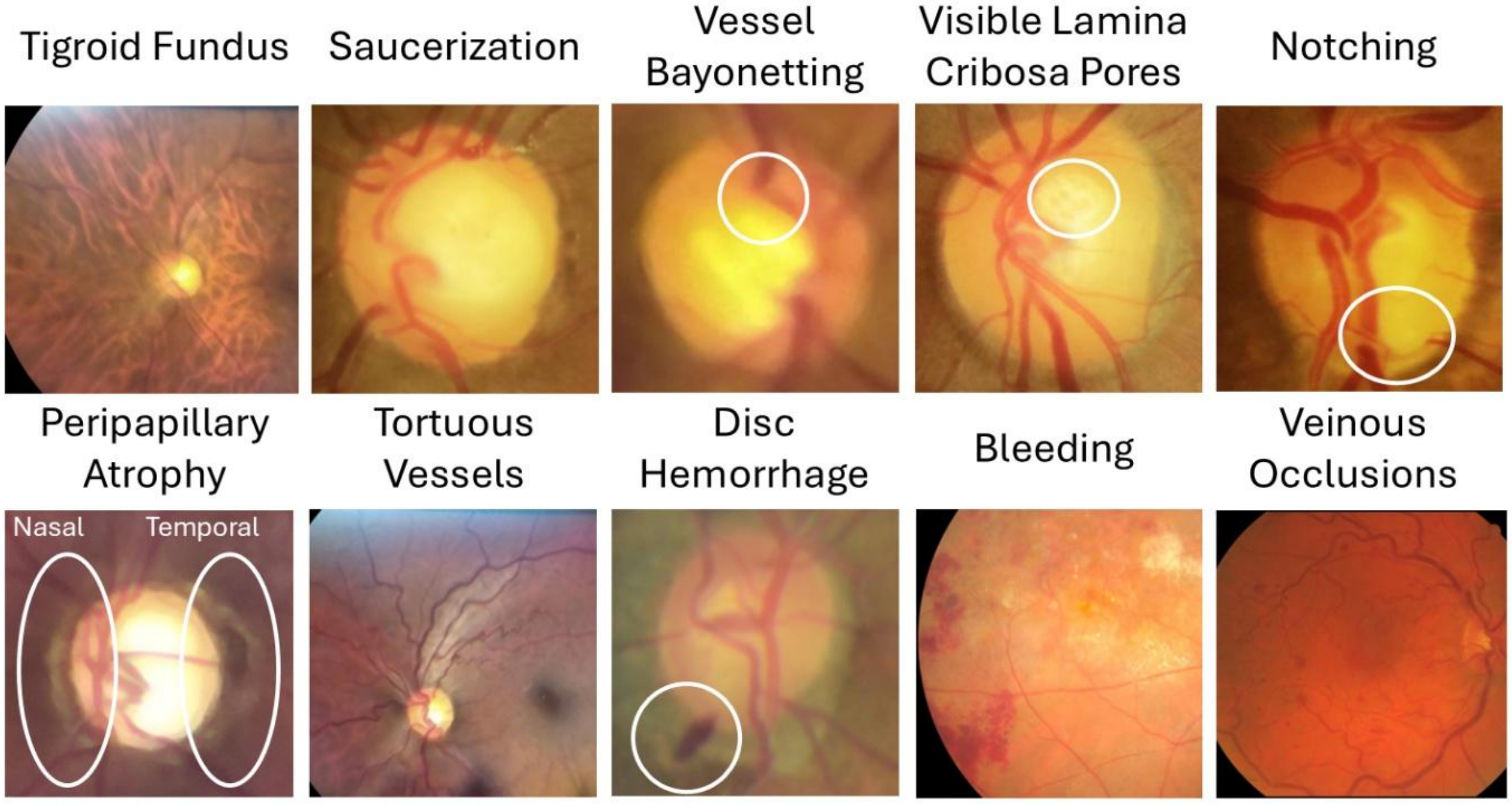

**Figure 2.** Examples of each of the eleven ONH features that were detected in the fundus photographs. Nasal and temporal peripapillary atrophy were assessed separately in photographs. Arterial occlusion was not detected in any photograph in the dataset.

**A. Image Analysis and Feature Extraction**

Fundus Photograph

Vessel Segmentation Model — UNET++ — Vessel Mask

Cup / Disc Segmentation Model — UNET++ — Cup/ Disc Mask — vCDR, hCDR; No Visible Disc

Feature Extractor — CNN Classifier — Ungradable; Detected Features

Discard

**B. Classification Model and Reporting**

ResNet18

Convolution Branch

Age

vCDR, hCDR

Detected Features

Tabular Branch

3-Class Classification Head

Glaucoma

Glaucoma Suspect

Non-Glaucoma

Interpretable Report

**Figure 3.** Overview of AI Pipeline. **A.** A single fundus photograph is processed by three component models: One, a UNET++ model to segment the vessels; two, a UNET++ model to segment the optic cup and disc, from which the vertical and horizontal cup-to-disc ratios (vCDR and hCDR) are derived; and three, a CNN-based feature detector to identify clinically relevant optic nerve head features and flag ungradable images for exclusion. **B.** The original fundus photograph, vessel and cup and disc masks, vCDR, hCDR, detected features, and participant's age are subsequently integrated into a model which classifies the eye as non-glaucoma, glaucoma suspect, or glaucoma, and generates a report to support clinical assessment.

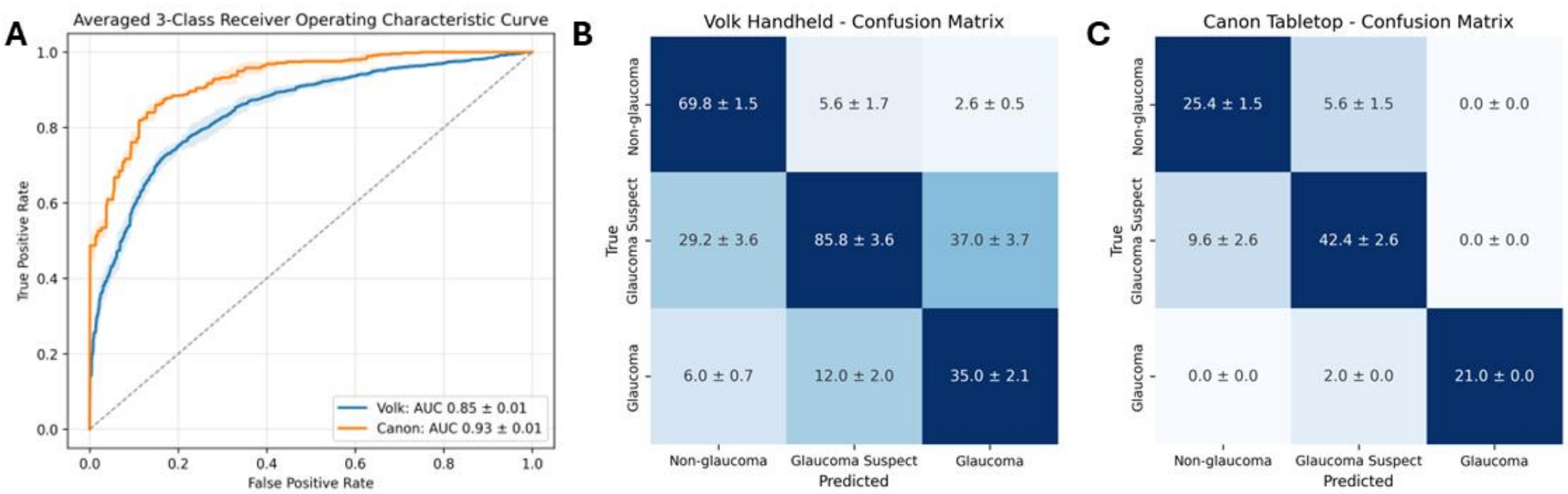

**Figure 4A.** Receiver Operating Characteristic Curve for both classification models. **B.** Volk Viva confusion matrix from mean of confusion matrix across the five folds. **C.** Canon CR-2-AF confusion matrix from mean of confusion matrix across the five folds.

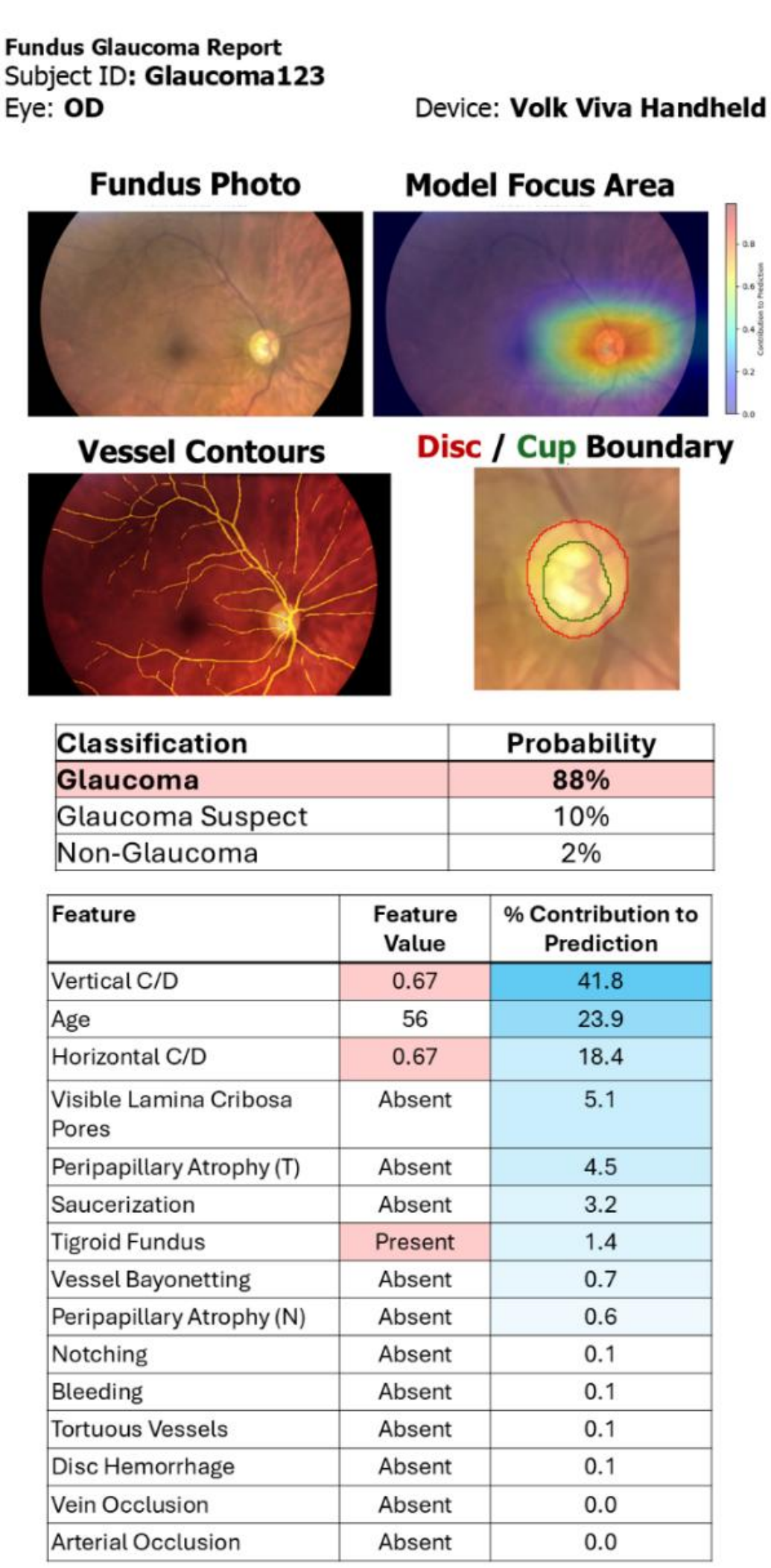

| Classification | Probability |
|---|---|
| Glaucoma | 88% |
| Glaucoma Suspect | 10% |
| Non-Glaucoma | 2% |

| Feature | Feature Value | % Contribution to Prediction |
|---|---|---|
| Vertical C/D | 0.67 | 41.8 |
| Age | 56 | 23.9 |
| Horizontal C/D | 0.67 | 18.4 |
| Visible Lamina Cribosa Pores | Absent | 5.1 |
| Peripapillary Atrophy (T) | Absent | 4.5 |
| Saucerization | Absent | 3.2 |
| Tigroid Fundus | Present | 1.4 |
| Vessel Bayonetting | Absent | 0.7 |
| Peripapillary Atrophy (N) | Absent | 0.6 |
| Notching | Absent | 0.1 |
| Bleeding | Absent | 0.1 |
| Tortuous Vessels | Absent | 0.1 |
| Disc Hemorrhage | Absent | 0.1 |
| Vein Occlusion | Absent | 0.0 |
| Arterial Occlusion | Absent | 0.0 |

**Figure 5.** An example report from a glaucomatous eye presents the original fundus photograph, the model focus area identifying the regions contributing most to the classification decision, it also overlays the segmented vessel contours on the photograph and outlines the disc and cup boundary identified. Below is a table presenting the class-specific confidence scores indicating the probability that the photograph should be classified as each class. Finally, the bottom of the report displays a table with the feature value extracted from each image by the feature detection model, and for each feature, how much it contributed to the prediction, relative to the other features.

# SUPPLEMENTARY MATERIAL

**Supplementary Table 1.** Ablation Test of Inputs

| Input removed | Canon CR-2-AF % drop in macro-AUC | Volk Viva % drop in macro-AUC |
|---|---|---|
| RGB image | 0.3% | 7.2% |
| Masks | 19.5% | 0.3% |
| Tabular features | 4.7% | 8.5% |

**Supplementary Table 2.** Shapley Additive exPlanation (SHAP) scores of tabular features

| Feature | Volk Viva (mean ± SD) | Canon CR-2-AF (mean ± SD) |
|---|---|---|
| Temporal peripapillary atrophy | 0.315 ± 0.025 | 0.211 ± 0.075 |
| Vertical Cup-to-Disc Ratio | 0.098 ± 0.046 | 0.065 ± 0.028 |
| Horizontal Cup-to-Disc Ratio | 0.088 ± 0.025 | 0.081 ± 0.021 |
| Age | 0.062 ± 0.022 | 0.070 ± 0.019 |
| Visible Lamina Cribosa Pores | 0.059 ± 0.013 | 0.015 ± 0.005 |
| Saucerization | 0.027 ± 0.003 | 0.020 ± 0.007 |
| Tigroid Fundus | 0.020 ± 0.008 | 0.000 ± 0.000 |
| Tortuous Vessels | 0.011 ± 0.004 | 0.134 ± 0.016 |
| Nasal Peripapillary Atrophy | 0.000 ± 0.000 | 0.011 ± 0.007 |
| Vessel Bayonetting | 0.000 ± 0.000 | 0.000 ± 0.000 |
| Disc Haemorrhage | 0.000 ± 0.000 | 0.000 ± 0.000 |
| Notching | 0.000 ± 0.000 | 0.000 ± 0.000 |
| Arterial occlusion | 0.000 ± 0.000 | 0.000 ± 0.000 |

| Bleeding | 0.000 ± 0.000 | 0.000 ± 0.000 |
|---|---|---|
| Veinous occlusion | 0.000 ± 0.000 | 0.000 ± 0.000 |

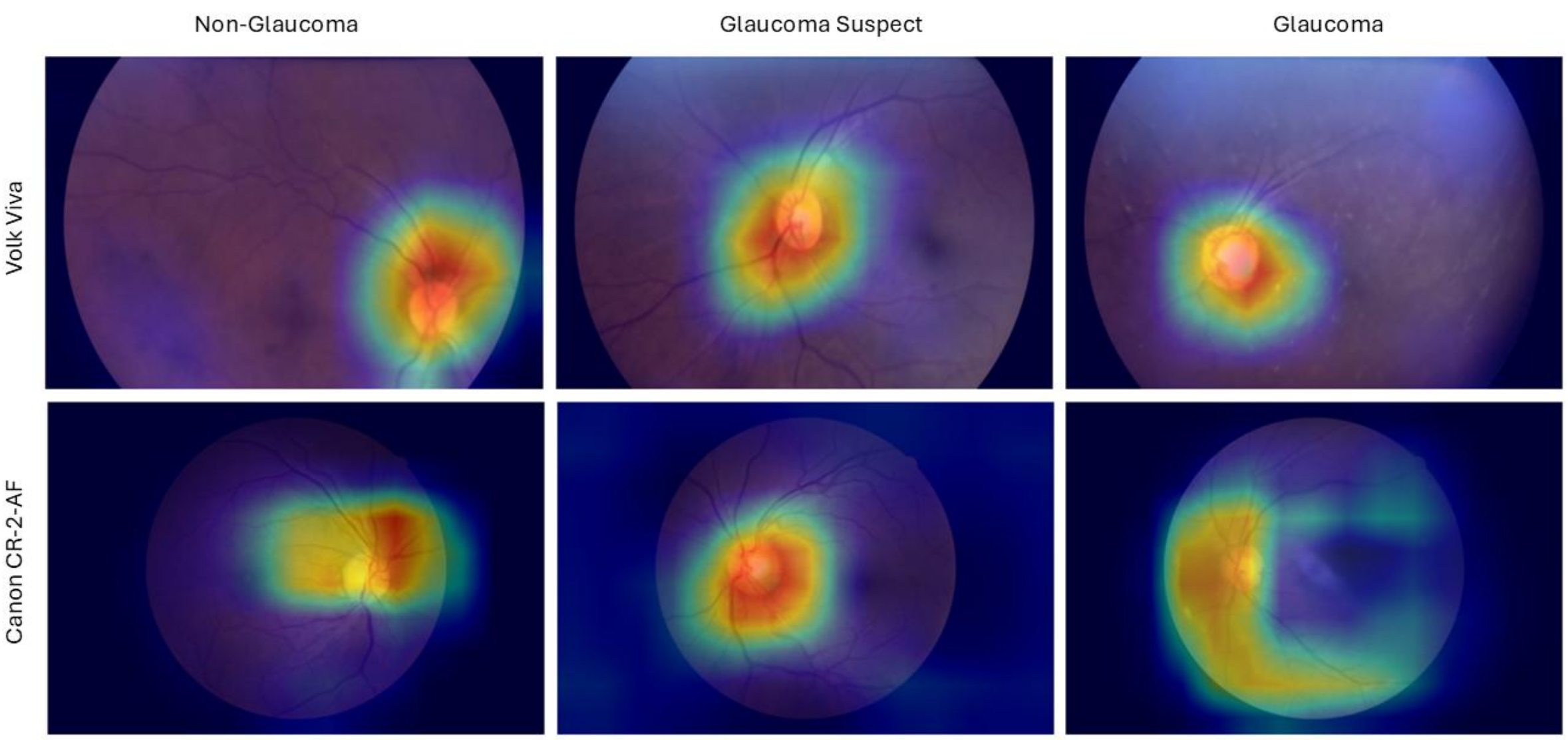

**Supplementary Figure 1.** Grad-CAM outputs for each image from each camera and diagnosis class revealed that model focus was primarily on the disc and cup region.